\documentclass{article}

\usepackage[final]{neurips_2023}

\usepackage[utf8]{inputenc} 
\usepackage[T1]{fontenc}    
\usepackage{hyperref}       
\usepackage{url}            
\usepackage{booktabs}       
\usepackage{amsfonts}       
\usepackage{nicefrac}       
\usepackage{microtype}      
\usepackage{xcolor}         

\usepackage{cite}           
\usepackage{multirow}
\usepackage{graphicx}
\usepackage{amsmath}

\usepackage{algorithm}      
\usepackage{algorithmic}

\title{
  Ascend HiFloat8 Format for Deep Learning
  }

\author{%
  Yuanyong Luo\thanks{Corresponding Author: luoyuanyong@\{hisilicon.com, yeah.net\}} , %
  Zhongxing Zhang, 
  Richard Wu,
  Hu Liu,
  Ying Jin\\
  \large{HiSilicon, Huawei}
  \\
  \And 
  Kai Zheng,
  Minmin Wang,
  Zhanying He\\
  \large{Central Hardware, Huawei}
  \\
  \And 
  Guipeng Hu,
  Luyao Chen,
  Tianchi Hu,
  Junsong Wang\\
  \large{Computing Product Line, Huawei}
  \\
  \And 
  Minqi Chen,
  Mikhaylov Dmitry,
  Korviakov Vladimir,
  Bobrin Maxim, 
  Yuhao Hu,
  Guanfu Chen,
  Zeyi Huang\\
  \large{Central Media, Huawei}
}

\begin{document}

\maketitle

\begin{abstract}

This preliminary white paper proposes a novel 8-bit floating-point data format HiFloat8 
(abbreviated as HiF8) for deep learning. 
HiF8 features tapered precision. For normal value encoding, it provides 7 exponent values with 
3-bit mantissa, 8 exponent values with 2-bit mantissa, and 16 exponent values with 1-bit mantissa. 
For denormal value encoding, it extends the dynamic range by 7 extra powers 
of 2, from 31 to 38 binades (notice that FP16 covers 40 binades). 
Meanwhile, HiF8 encodes all the special values except that positive zero and 
negative zero are represented by only one bit-pattern.
Thanks to the better balance between precision and dynamic range, HiF8 can be 
simultaneously used in both forward and backward passes of AI training. 
In this paper, we will describe the definition and rounding methods of HiF8, as well as 
the tentative training and inference solutions. 
To demonstrate the efficacy of HiF8, massive simulation results on various neural 
networks, including traditional neural networks and large language models (LLMs), will 
also be presented.  

\end{abstract}

\section{Introduction}

In 2020, several of our top chip architects foresaw that as Moore's Law \cite{Golio2015} 
slows down, low-precision training and inference would be an important way to reduce the 
computing power and mitigate the memory wall \cite{Kwon2018} for AI hardware. 
Thus in 2021, HiSilicon launched the HiFloat project, 
aiming to study and develop novel low-precision data formats for our AI products. 
Subsequently, this project attracted many researchers from other departments to join. 
In this paper, we will officially disclose some of our research achievements on Float8. 

First, let's take a brief look at history. 
Generally, the development of AI data formats can be divided into the 
following four phases based on the time for commercial use:

\textbf{Phase 1 (1959 - 2006): FP64.}
The concept of AI was first proposed in 1956. 
Then, several theoretical prototypes and directions were established and developed, 
including backpropagation algorithm \cite{Linnainmaa1970}, convolutional neural network (CNN)
\cite{LeCun1995}, and long short-term memory \cite{Hochreiter1997}. 
During this period, CPUs were the major AI hardware, and the mainstream data format was the 
64-bit double precision floating-point format FP64 defined in IEEE Standard for Floating-Point 
Arithmetic \cite{IEEE754-1985} in 1985. 
Prior to this, some predecessors of FP64 were in use. 

\textbf{Phase 2 (2006 - now): FP32.}
In 2006, FP32 was used for the first time to train CNNs on GPUs and achieved a four-fold 
performance improvement compared with FP64 training on CPUs \cite{Chellapilla2006}. 
The bit width of FP32 is only half that of FP64. 
Therefore, a chip that uses the FP32 format can store more data and integrate more 
multiply-accumulate (MAC) units than one that uses FP64. 
The single instruction, multiple data (SIMD) computing mode enables GPUs to have a significantly 
higher degree of computing parallelism than CPUs \cite{Nuzman2006}. 
Finally, AlexNet \cite{Krizhevsky2012} which ran on two GPUs won the championship of ILSVRC 2012, 
making FP32 the mainstream data format for deep learning training from 2012 to 2017.

\textbf{Phase 3 (2017 - now): Float16 mixed precision.}
In 2016, Google proposed a novel Float16 data format in its TensorFlow white paper \cite{Abadi2016}. 
This format was equivalent to FP32 without the last 16 bits and was used as the input
data format of multiplication, whereas FP32 was used for accumulation in the general 
matrix-matrix multiplication (GEMM). 
This format was later named Brain Floating Point 16 (BF16) \cite{Kalamkar2019} and deployed on 
Tensor Processing Unit (TPU) \cite{Jouppi2017} V2 and V3 .
In 2017, Google's AlphaGo \cite{Silver2017} powered by TPU V2 defeated the world's number one Go 
player Ke Jie in all three games, causing a global sensation. 
Unlike Google, NVIDIA and Huawei adopt the IEEE 754 half-precision data format \cite{Zuras2008}.
NVIDIA released the V100 GPU \cite{Markidis2018} in 2017 and Huawei shipped the Ascend 
NPU \cite{Liao2019} in 2019. 
Both the GPU and NPU support the FP16 and FP32 mixed precision training strategy, with 
backward global loss-scaling to prevent excessive zero-valued activation gradients caused 
by narrow dynamic range of FP16 \cite{Micikevicius2017}. 
Since 2017, the Float16 mixed precision training solution, mainly consisting of BF16 and FP16, 
has become the main choices for deep learning. 
However, FP32 is still used for some networks that require high precision.

\textbf{Phase 4 (2022 - now): Float8 mixed precision.}
In 2018, IBM radically cut off the last 8 bits of FP16 to obtain a Float8 data format 
with 1 sign bit, 5 exponent bits, and 2 mantissa bits (E5M2) \cite{Wang2018}. 
However, on MobileNetV2 and Transformer networks, the accuracy of E5M2 training
decreased dramatically. 
Then in 2019, IBM further proposed the 8-bit hybrid FP8 (HFP8) training solution, with E4M3 
(a Float8 format that has 1 sign bit, 4 exponent bits, and 3 mantissa bits) as the weights 
and activations format and E5M2 as the activation gradients format \cite{Sun2019}. 
Thus, when calculating gradients in the backward pass, GEMM that supports mixed Float8 
inputs is needed, and HFP8 is therefore called FP9 in hardware implementation \cite{Agrawal2021}. 
In 2022, NVIDIA deployed HFP8 (renamed as FP8) mixed precision solution with twice the computing 
power of FP16 on its new H100 GPU, and shipped by the end of the year \cite{Elster2022}.
Subsequently, Intel, ARM, AMD, and Quanlcomm et al., announced that they would also support this 
solution \cite{Micikevicius2022,Rouhani2023}.

From a brief historical review, we can see that low-precision training \cite{SabbaghMolahosseini2012} 
has always been an important direction to improve AI performance. 
And the commercial use of Float8 mixed precision has begun. 
However, it is also important to note that the trade-off between precision and dynamic range is 
challenging in the evolution from Float16 to Float8. 
For Float16, one fixed field-width format (either BF16 or FP16), could be elegant to cover all 
GEMM inputs, and works well for almost all scenarios. 
But this failed in Float8 mixed precision.
In addtion to the research for fixed field-width formats, Posit \cite{Gustafson2017} 
is a decent exploration for tapered precision data type, because it matches the centralized 
characteristic of AI data distribution during training and inference. 
Unfortunately, Posit16 failed in the competition for Float16 mixed precision, due to its larger 
hardware cost than BF16 and FP16, and insignificant precision improvement for training \cite{Lu2019}. 
And Posit8 also failed in the competition for Float8 mixed precision, because its encoding method 
cannot balance the precision and dynamic range very well to meet the training 
requirement \cite{Lu2020,Raposo2021}. 

Inspired by FP16, Posit, and HFP8/FP8, this paper proposes a novel 8-bit floating point format 
HiF8 for deep learning, which features the better balance between precision and dynamic range 
compared with the existing Float8 formats, and can be simultaneously used in both forward and 
backward passes for AI training. 
A large number of simulation results would be presented to illustrate the advantages of HiF8 
in this paper.

\section{HiFloat8}

This section first describes the definition of the novel 8-bit floating-point data format HiF8, 
including the support for special values. 
Then, some consideration and design issues for HiF8 will be explained. 

\subsection{Novel Data Format}

We propose a new general-purpose floating-point enconding and decoding method for data 
expression, for which the field width, dynamic range, and significand precision can be 
scaled based on scenario requirements. 
This paper focuses on the 8-bit floating-point instance for deep learning usage. 
On the basis of the IEEE 754 \cite{Zuras2008}, HiF8 defines an additional \emph{dot field}. 
Therefore, HiF8 consists of the four fields as listed in Table~\ref{HiF8-Fields}: 
a sign field, a dot field, an exponent field, and a mantissa field. 

\begin{table}[htbp]
  \caption{Fields of HiF8}
  \label{HiF8-Fields}
  \centering
  \begin{tabular}{cclll}
    \toprule
                    & Sign   & Dot: D          & Exponent: E        & Mantissa            \\
    \midrule
                    & 1      & 2: \{2, 3, 4\}  & D: $\pm [2, 15]$   & 5 - D = [1, 3]      \\
      Width: Values & 1      & 3: 1            & D: $\pm 1$         & 4 - D = 3           \\
                    & 1      & 4: 0            & D: 0               & 3 - D = 3           \\
                    & 1      & \multicolumn{2}{l}{4: DML --- Denormal Sign}     & 3       \\
    \bottomrule
  \end{tabular}
\end{table}

The following describes each field in detail:
\begin{itemize}

  \item \emph{Sign Filed: 1 bit}, determining the sign of the HiF8 number, which is 
  the sign of the significand as well. 
  By default, 1 indicates the negative sign and 0 indicates the positive sign. 

  \item \emph{Dot Field: 2 to 4 bits}, used to code the five D values (0 to 4) and 
  the sign of DenorMaL (DML). 
  D value explicitly indicates the number of bits occupied by the exponent field, and 
  implies the number of bits occupied by the mantissa field. 
  DML sign specifies that the HiF8 number has no exponent field, and needs to 
  be parsed by denormal equation (\ref{DML-Eq}). 
  Otherwise, NorMaL (NML) equation (\ref{NML-Eq}) should be used. 
  The dot field is coded using unconventional prefix codes, that is, the 4-bit 
  width is used for coding small value 0 and DML sign, the 3-bit width is used for 
  coding medium value 1, whereas the 2-bit width is used for coding large values 2, 
  3, and 4. 
  Table \ref{Dot-Encoding} specifies the default mapping, in which values with a 2 
  in the subscript are binary, otherwise they are decimal. 

  \begin{table}[htbp]
    \caption{Unconventional Prefix Code for Dot Field}
    \label{Dot-Encoding}
    \centering
    \begin{tabular}{lcccccc}
      \toprule
      Width    & \multicolumn{3}{c}{2}    & 3          & \multicolumn{2}{c}{4}                \\
      \midrule
      Code     & $11_2$     & $10_2$      & $01_2$     & $001_2$    & $0001_2$     & $0000_2$ \\
      Value    & 4          & 3           & 2          & 1          & 0            & DML      \\
      \bottomrule
    \end{tabular}
  \end{table}

  \item \emph{Exponent Field: D bits} (an implicit leading magnitude bit with value 
  1 unstored), where D is equal to the coded value of the dot field 
  and $D \in \{0, 1, 2, 3, 4\}$. 
  Different from the offset-binary method used in the IEEE 754, the exponent field 
  of HiF8 uses \emph{sign-magnitude} code to represent values. 
  But the most significant bit (MSB) of the magnitude is fixed to 1. 
  Mark the sign of exponent as Se. 
  By default, 1 means negative sign and 0 means positive sign. 
  Then denote the complete sign-maginitude exponent as Ei in binary mode, we have:
  \[ Ei = \{Se, Mag[1 : end]\} = \{Se, 1, Mag[2:end]\} \]
  Since the fixed MSB of the magnitude can be hidden, Ei is simplified as Em in 
  the memory format: 
  \[ Em = \{Se, Mag[2:end]\} \]
  The number of bits in Em equals the value of D.
  When D is zero, Em dose not occupy any bit width, indicating that the exponent 
  value equals zero. 

  \item \emph{Mantissa Field: 1 to 3 bits} (an implicit leading significand bit 
  with value 1 unstored), coded by unsigned integer. 
  For the normal number of HiF8, mantissa represents the fractional bits 
  (to the right of the binary point) in the significand. 
  For the denormal value of HiF8, mantissa represents the extended exponents in a 
  biased form. 

\end{itemize}

So far, we have introduced the four fields for HiF8. 
Table \ref{HiF8-Mapping} summarizes the corresponding code-value mapping details. 
In the memory format, dot field is stored as outlined in Table \ref{Dot-Encoding}. 
Exponent field is stored as Em with an implicit bit.
Em can be further interpreted as Ei in binary mode, and E in decimal mode. 

\begin{table}[htbp]
  \caption{HiF8 Code-Value Mapping Details}
  \label{HiF8-Mapping}
  \centering
  \begin{tabular}{lccllll}
    \toprule
    Value of Dot   & DML  & 0  & 1  & 2  & 3  & 4 \\
    \midrule
    Em (binary)    & N/A  & None  & Se  & Se, Mag[2]  & Se, Mag[2:3]  & Se, Mag[2:4] \\
    Ei (binary)    & N/A  & 0  & Se, 1  & Se, 1, Mag[2]  & Se, 1, Mag[2:3]  & Se, 1, Mag[2:4] \\
    E (decimal)    & N/A  & 0  & $\pm 1$  & $\pm [2, 3]$  & $\pm [4, 7]$  & $\pm [8, 15]$ \\
    Mantissa Width & 3  & 3  & 3  & 3  & 2  & 1 \\
    \bottomrule
  \end{tabular}
\end{table}

Denote the sign as S, and the mantissa as M. 
For the normal number, HiF8 should be interpreted as: 
\begin{equation}
  X = (-1)^{S} \times 2^{E} \times 1.M       \label{NML-Eq}
\end{equation}  
In the normal equation (\ref{NML-Eq}), 2 bit-patterns with the largest absolute value 
($2^{15} \times 1.5$) should be interpreted specially for \emph{Infinities}. 
For the denormal number, HiF8 should be interpreted as:
\begin{equation}
  X = (-1)^{S} \times 2^{M - 23} \times 1.0  \label{DML-Eq}
\end{equation}
In the denormal equation (\ref{DML-Eq}), $M \in [1, 7]$, which offers 7 addtional 
exponent values of [-22, -16]. 
2 bit-patterns with M = 0 should be interpreted specially for 
\emph{Zero and NaN} (Not a Number). 

\begin{table}[htbp]
  \caption{Typical Values and Features for HiF8, FP8 and FP16}
  \label{HiF8-FP8-Coding}
  \centering
  \begin{tabular}{lllll}
    \toprule
                  & FP16  & HiF8 & FP8-E4M3 & FP8-E5M2 \\
    \midrule   
    Infinities       & Support   & $ S{\color{red}11} {\color{green}0111} {\color{blue}1}_2 $ & N/A & Support \\
    NaN              & Support   & $ 1{\color{red}0000} {\color{blue}000}_2 $ & Support & Support  \\
    Zero             & Support   & $ 0{\color{red}0000} {\color{blue}000}_2 $ & Support & Support  \\
    Max Positive NML & $2^{15}\times(2-2^{-10}) $ & $ 0{\color{red}11} {\color{green}0111} {\color{blue}0}_2 = 2^{15}  $ & $ 1.75 \times 2^8 $ & $ 1.75 \times 2^{15} $ \\
    Min Positive NML & $2^{-14}$ & $ 0{\color{red}11} {\color{green}1111} {\color{blue}0}_2 = 2^{-15} $ & $ 2^{-6} $ & $ 2^{-14} $ \\
    Max Positive DML & $2^{-14}\times(1-2^{-10}) $ & $ 0{\color{red}0000} {\color{blue}111}_2 = 2^{-16} $ & $ 1.75 \times 2^{-7} $ & $ 1.5 \times 2^{-16} $ \\
    Min Positive DML & $2^{-24}$ & $ 0{\color{red}0000} {\color{blue}001}_2 = 2^{-22} $ & $ 2^{-9} $ & $ 2^{-16} $ \\
    Exponent         & [-14, 15] & [-15, 15] & [-6, 8] & [-14, 15] \\
    Exponent (+DML)  & [-24, 15] & [-22, 15] & [-9, 8] & [-16, 15] \\
    \bottomrule
  \end{tabular}
\end{table}

\begin{figure}[htbp]
  \centering
  \includegraphics[width=13cm]{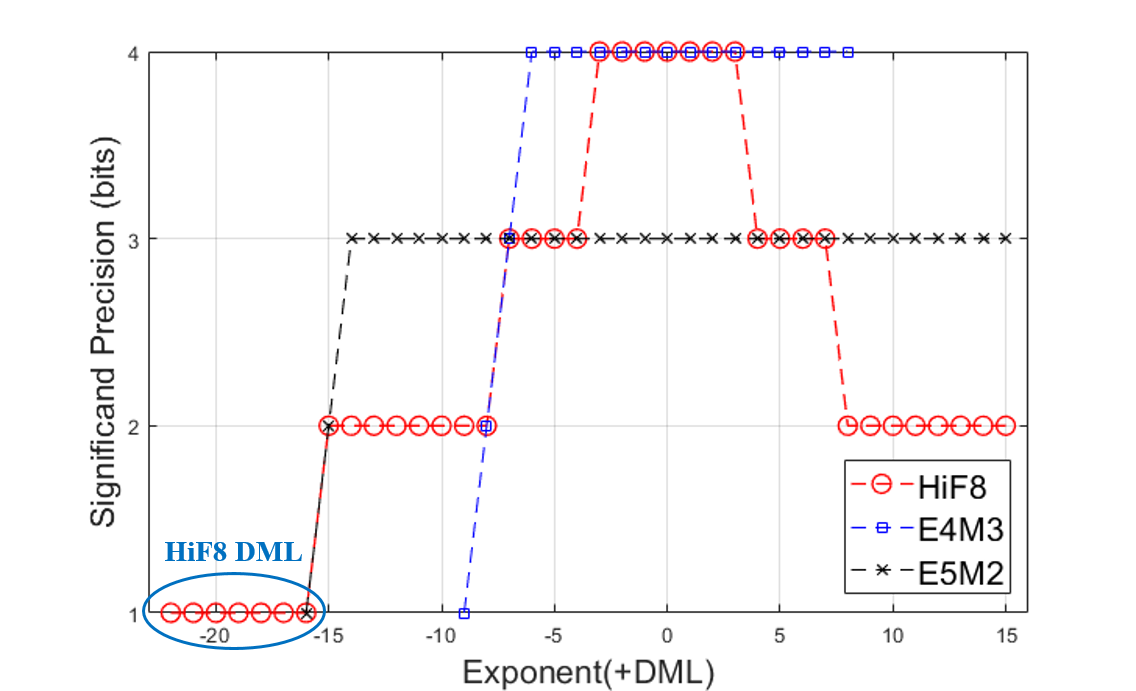}
  \caption{Significand Precision over Exponent}
  \label{Sig-Exp}
\end{figure}

Table \ref{HiF8-FP8-Coding} shows some typical encoding values and features for HiF8, FP8 
\cite{Micikevicius2022} and FP16 \cite{Zuras2008} formats. 
For the HiF8 format in binary mode, the black bit is the sign field, the red bits are the 
dot field, the green bits are exponent field, and the blue bits are the mantissa field. 
Obviously, HiF8 supports all special values, but does not distinguish between positive zero 
and negative zero because it's not necessary for deep learning. 
In addition to the regular exponent range, Table \ref{HiF8-FP8-Coding} also lists the 
exponent range after denormal to normal operation, which is critical to measuring the 
dynamic range of a floating point format. 
To clearly illustrate the differences between HiF8 and FP8, Fig. \ref{Sig-Exp} further 
plots the distribution of significand bits on the exponent after denormal to normal operation.

\subsection{Consideration and Design}

Before the launch of the HiFloat project, we investigated lots of literatures on
low-percision training. 
Finally, three data formats were selected as the references. 
The first is the FP16 defined by the IEEE 754 \cite{Zuras2008}, with the global 
backward loss-scaling method \cite{Narang2017}, this data type trains well for 
almost all neural networks. 
The second is the HFP8 \cite{Sun2019} proposed by IBM, which is the first 8-bit 
floating point format for commercial use \cite{Elster2022}. 
The third is the Posit \cite{Gustafson2017} invented by John Gustafson, 
which features the tapered precision that matches the centralized characteristic 
of AI data distribution. 
In the following, we will explain several design considerations for HiF8 and show 
how those references guide our design. 

\begin{itemize}
  
  \item \emph{Dot field}
   
  \textbf{Consideration}: To avoid multi-float8 formats like HFP8 for deep learning, 
  tapered precision is a promissing direction to explore. 
  Posit realizes the tapered precision by adopting a variable-length field 
  called regime to encode two base values of the exponents. 
  However, the regime field using unary coding is not convenient and flexible 
  enough to manipulate the significant bits distribution over the exponent to  
  better match the training requirements. 

  \textbf{Design}: Therefore, HiF8 adopts the flexible prefix code to form a novel 
  dot field, which directly indicates the width of exponent and the sign of denormal. 
  Moreover, to smooth the precision variation and avoid the mantissa width jumping 
  down by more than 1 bit, we use large width to code small numbers and small width 
  to code large numbers. 
  One can obtain an in-depth analysis from Table \ref{HiF8-Fields} and 
  Table \ref{Dot-Encoding}. 
  
  \item \emph{Exponent field}
  
  \textbf{Consideration}: To avoid coding redundancy, we must ensure that the exponent 
  values pointed by each value of the dot field do not overlap with each other. 
  Coding methods like the offset-bianry used in FP16 and the special one used in Posit, 
  cannot achieve this goal. 
  Fortunately, there are other three signed number reprentations for us to consider: 
  sign-magnitude, one's complement, and two's complement. 

  \textbf{Design}: Inspired by the implicit bit design of the mantissa field in FP16, 
  HiF8 chooses the sign-magnitude coding for the exponent field, where one implicit bit 
  is fixed to 1 to avoid repetitive representation of the exponent values. 
  As shown in Table \ref{HiF8-Mapping}, HiF8 becomes non-redundant in data expression. 

  \item \emph{Denormal Mode}

  \textbf{Consideration}: Without the denormal design, HiF8 would have a 4-bit 
  mantissa when the exponent is equal to zero, but would only support 31 exponent 
  values from -15 to 15. 
  Currently, the activation gradients of LLMs require a higher dynamic range of the 
  data format \cite{Perez2023}. 
  While from the practice of HFP8 or FP8, we can conclude that 3-bit mantissa is 
  sufficient for training tasks. 
  Thus a better balance is possible to extend the dynamic range of HiF8. 

  \textbf{Design}: We reduce the mantissa width from 4 bits to 3 bits when the exponent 
  is equal to zero. 
  Then the resulting free coding spaces are directly used to expand the exponent range, 
  as formulated by the denormal euqation (\ref{DML-Eq}) and depicted in 
  Fig. \ref{Sig-Exp}. 
  In this way, the binades that HiF8 can cover increase from 31 to 38, very close to 
  the 40 of FP16. 
  
\end{itemize}

From the above analyses, we can see that HiF8 stands on the shoulders of the three 
data formats, absorbs their advantages, and finally achieves a better balance for 
deep learning at the 8-bit limit. 
Compared with FP8, HiF8 takes both precision and dynamic range into account, and 
is capable of replacing two formats of FP8 with only one format. 
Compared with Posit(8, 2) \cite{Lu2020, Raposo2021}, HiF8 has 31 exponents with 
mantissa no less than 1 bit, while Posit(8, 2) only has 24 exponents with mantissa 
no less than 1 bit. 
And compared with FP16, HiF8 has almost the same dynamic range, which is much better 
than FP8-E4M3 and FP8-E5M2.

\section{Rounding Methods}

In Float8 mixed precision training and inference, high-precision floating-point formats 
such as FP32 need to be converted into low-precision format Float8, and then input to 
GEMM, during which rounding is involved. 
As the precision of Float8 is relatively lower than BF16 and FP16, the rounding method 
is extremely sensitive to the convergence and accuracy of neural network training. 
After the theoretical analysis and a large number of experiments, we conculde that HiF8 
will support two rounding methds:  rounding half to away (from zero), and hybrid 
rounding. 
To covert high-precision data to HiF8, we use only rounding half to away in the 
forward pass, and rounding half to away or hybrid rounding in the backward pass. 
In addition, to meet the requirements of certain AI algorithms, HiF8 also provides two 
options: saturation to boundary upon overflow, and NaN saturation to zero. 
The following describes the rounding methods during the conversion from high-preciosn 
formats to HiF8. 

\subsection{Rounding Half}

Rounding half (rounding to nearest) yields an error of 0.5 ulp (unit of least precision), 
and can be generally classfied into rounding half to even (TE) and rounding half to away 
(TA) \cite{Zuras2008}. 
Technically, if the MSB of the discarded bits is 1 and the other discarded bits are 
all 0, TE would ensure that the LSB (least significant bit) of the rounded number 
is even by carrying or not. 
Otherwise, both TE and TA would carry as long as the MSB of the discarded bits is 1. 
Although TA features easier hardware implementation, TE is used by default in most 
papers and commercial products, because it maximizes the unbiasedness 
\cite{Micikevicius2022}. 
In fact, the occurrence probability of the TE special case is extremely low during 
the conversion from high-precision formats to HiF8. 
For example, if HiF8 reserves 3-bit mantissa, the probability is only $2^{-20}$ 
during the conversion from FP32 to HiF8. 

The biggest challenge of Float8 in AI is its limited data resolution capability. 
The analysis result shows that the data resolution capability of TA is slightly 
higher than that of TE. 
Consider a TE special case of three 3-bit numbers with consecutive integer 
bits: 00.1, 01.1, and 10.1. 
TE is rounded as TE(00.1) = 00, TE(01.1) = 10, TE(10.1) = 10, giving two 
different results. 
TA is rounded as TA(00.1) = 01, TA(01.1) = 10, TA(10.1) = 11, giving three 
different results. 
Therefore, in the TE special case, TA enables a higher data resolution 
capability of Float8 than TE. 
The simulation experiments of HiF8 also evidence that TA produces slightly higher 
training accuracy than TE. 
For example, the TA-based training accuracies of ResNet50 \cite{He2016} and 
MobileNet\_V2 \cite{Sandler2018} are 0.06\% and 0.11\% higher than the TE-based 
training accuracies on average. 

Since TA features simpler hardware implementation and higher training 
accuracy, it is therefore supported during the conversion from 
high-precision formats (including FP32, FP16, and BF16) to HiF8. 

\subsection{Hybrid Rounding}

Large-scale HiF8 mixed precision training experiments show that global TA 
rounding works well for almost all neural networks. 
But for YoLo-V3-Tiny \cite{Redmon2018}, some segments of the loss curve crashed. 
As a result, the final accuracy was 1.67\% lower than the FP32 baseline. 
After extensive research and many experiments, in addition to the global 
TA rounding method, we propose a second HiF8 rounding method for training, 
which combines TA rounding for the forward pass and hybrid rounding for 
the backward pass. 
This makes the training accuracy of the YoLo-V3-Tiny close to the baseline value.
Therefore, HiF8 supports both TA rounding and hybrid rounding (HR).
In fact, HR is essentially an optimized version of standard stochastic 
rounding \cite{Connolly2021}, which is easier to implement in circuits and has 
slightly better training accuracy. 

The error of stochastic rounding (SR) is 1 ulp. 
Compared with TA, SR has a significant advantage when data is processed in batches. 
Specifically, in SR, a uniformly distributed random number needs to be randomly 
generated and used as threshold T, $(T \in [0, 1))$.  
All bits to be discarded are regarded as fractional bits and marked as F, 
$(F \in [0, 1))$. 
If $F \ge T$, 1 is added to the reserved bits K, otherwise 0 is added to the reserved 
bits K. 
As threshold T is uniformly distributed, the expected value after SR is expressed as 
follows:  

\[
  (K + 1) \times F + K \times (1 - F) = K + F 
\]

Apparently, SR can maximize the invariance of the overall mean value during the 
rounding of batch data. 
However, deep learning needs to generate a large number of uniformly distributed 
random numbers in parallel, both software and hardware implementations of SR hit 
a performance bottleneck \cite{LEcuyer2017}. 

\begin{figure}[htbp]
  \centering
  \includegraphics[width=16cm]{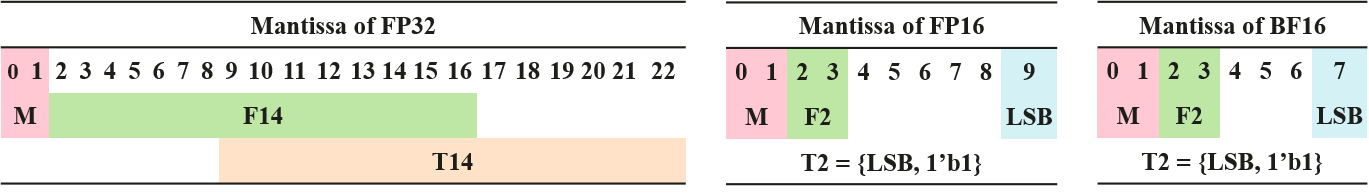}
  \caption{Threshold and Fraction of Simplified Stochastic Rounding}
  \label{SR-TF}
\end{figure}

To tackle the bottleneck, we first come up with a simplified SR hardware 
solution. 
Theoretical analysis and experiments show that the lower mantissa bits of 
floating-point numbers obey uniform distribution. 
Therefore, for the FP32 source data, we set 14 LSBs of the source format 
as the threshold T14, and 14 MSBs of the discarded bits as the fractional 
bits F14. 
However, for FP16 and BF16 source data, the discarded bits are not wide 
enough to be divided into reasonable and weakly relevant threshold and 
fraction. 
To solve this dilemma, we combine a fixed 1 and the LSB of the source 
format into a special 2-bit threshold T2, and set the 2-bit MSBs of the 
discarded bits as the fractional bits F2. 
As illustrated in Fig. \ref{SR-TF}, we have designed a 14-bit SR (SR14) 
for FP32 and a 2-bit SR (SR2) for FP16 and BF16, without generating the 
random numbers by complex algorithms.
SR14 is very similar to standard SR, with the same 1 ulp rounding 
error.
SR2 is weakly stochastic with only 2 thresholds of 0.25 and 0.75, 
but has a small rounding error of 0.75 ulp. 
So far, by comparing F14/F2 with T14/T2, simplified SR can be done in 
hardware. 

HiF8 training experiments show that the effect of simplified SR is quite 
close to that of standard SR, but there is still a very small gap.
In fact, simplified SR achieves better mean invariance than 
TA, but incurs greater rounding error. 
Meanwhile, from Fig. \ref{Sig-Exp} and the centralized characteristic 
of AI data, we can make the following logical inference.
Most of the data is within the high-precision range of HiF8, in which TA 
rounding will cause only a small change in the average value, becuase the 
rounding direction is relatively balanced for large data volume. 
However, a small amount of data (especially large values) is within the 
low-precision range of HiF8, in which TA rounding may incur a large 
change in the average value , because the rounding direction can be 
unbalanced for small data volume. 
Through such analysis, we propose a hybrid rounding method as follows. 

\begin{equation}
  Y =  
  \begin{cases}
    \text{TA Rounding}   & \text{if } |E| < 4, \\
    \text{Simplified SR} & \text{if } |E| \ge 4.    
  \end{cases}     \label{HR}
\end{equation}  

In the HR rounding equation (\ref{HR}), E is the exponent value of 
the source data format. 
Experiments show that HR yields slightly better training accuray than 
the standard SR with low hardware cost. 
Especailly, for YoLo-V3-Tiny training, the baseline accuracy 
of FP16 mixed precsion is 16.63\%. 
When using global TA rounding, the training accuracy of HiF8 is 14.96\%, 
1.67\% lower than the baseline. 
When using TA rounding for the forward pass and standard SR for the 
backward pass, the training accuracy of HiF8 is 16.43\%, 0.20\% lower 
than the baseline. 
When using TA rounding for the forward pass and hybrid rounding for 
the backward pass, the training accuracy of HiF8 is 16.69\%, slightly 
better than the baseline.

Thus, in addition to the TA rounding, the proposed hybrid rounding is also 
supported during the conversion from high-precision formats 
(including FP32, FP16, and BF16) to HiF8.
Note that there is no need to use hybrid rounding in the forward pass,  
because the distribution of activations and weights is relatively more 
concentrated than the distribution of gradients in the backward pass.
At the same time, for the vast majority of neural networks, training 
with TA and HR makes very little difference, and so far we have only 
found one neural network that definitely needs HR.

\section{Experiments on Traditional Neural Networks} 

Currently, no hardware platform is available to support HiF8 data 
type and complete the computation process. 
Thus both training and inference experiments of traditional neural 
networks and LLMs, were performed with simulated HiF8 format. 
Specifically, using the rounding methods described above, the tensor 
values were converted from high-precision formats to only those that 
could be represented in HiF8. 
Hardware platforms, including Huawei Ascend NPUs 
\cite{Liao2019,Wu2022} and NVDIA GPUs 
\cite{Markidis2018,Choquette2020}, and software framework PyTorch 
\cite{Paszke2019}, were utilized to conduct the HiF8 training and 
inference experiments. 

Since traditional neural networks and LLMs have been identified to 
have varying degrees of data dispersion, we handle them differently.
In this section, only empirical results for the traditional neural 
networks are presented.

\subsection{Training with Backward Loss-Scaling}
\label{BLS}

As mentioned earlier, the dynamic range of HiF8 is almost the same 
as that of FP16, so for the traditional neural network, we inherit 
the training method of FP16 mixed precision.
Specifically, only the GEMM inputs, including activation, weight, 
and activation gradient tensors, are changed from FP16 to HiF8 
(excluding the last fully-connected layer). 
The others such as non-linearities or normalizations, remain the 
same as the FP16 mixed precision training. 
Most importantly, backward global loss-scaling is enabled 
to avoid excessive zero-valued gradients, which is the core trick 
to make both FP16 and HiF8 training effective \cite{Micikevicius2017}. 
As for the conversion from high-precision formats to HiF8, only TA 
rounding is used in the forward pass, and either TA rounding or 
hybrid rounding is used in the backward pass. 

We trained FP16 baseline results and HiF8 results 
with the same model architectures, weight initializations (non-fixed random 
 seeds), and optimizer hyper-parameters, and compared them. 
To verify the training accuracy of HiF8 for the traditional nerural 
networks, we chose two main application directions for simulation 
experiments: computer vision and natural language processing (NLP). 
The subdivided application scenarios of computer vision include 
classification, detection, and segmentation. 
In this paper, the most widely used typical networks based on the CNN 
and Transformer structures are selected for the experiments, including 
ResNet series \cite{He2016}, ResNeXt \cite{Xie2017}, VGG \cite{Simonyan2014}, 
MobileNet \cite{Sandler2018}, Inception \cite{Szegedy2016}, 
EfficientNet \cite{Tan2019}, DenseNet \cite{Huang2017}, 
ViT series \cite{Dosovitskiy2020}, YoLo series \cite{Redmon2018}, and    
DeepLab \cite{Chen2017}. 
For NLP, the mainstream Transformer models \cite{Vaswani2017, Kenton2019} 
are used. 

\begin{figure}[htbp]
  \centering
  \includegraphics[width=17.75cm]{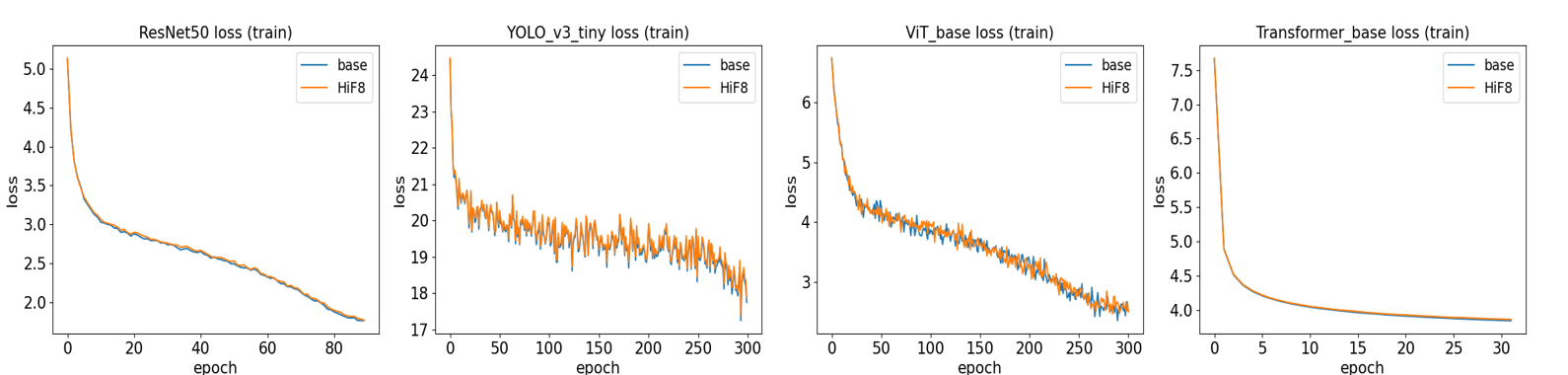}
  \caption{Training Loss Curves for Traditional Neural Networks}
  \label{TNN-Curve}
\end{figure}

Fig. \ref{TNN-Curve} compares the loss curves of HiF8 mixed precision 
training and FP16 mixed precision training (base) on the ResNet50, 
YoLo-V3-Tiny, ViT-Base, and Transformer-Base neural network models. 
The loss curves of HiF8 are highly overlapped with that of FP16, therefore 
the convergence speed of HiF8 is the same as that of FP16, and both HiF8 
training and FP16 training can be completed within the same number of epochs. 

\begin{table}[htbp]
  \caption{Validation Accuracy for Traditional Neural Networks}
  \label{TNN-Training}
  \centering
  \begin{tabular}{llclc}
    \toprule 
    Model & Metric & FP16 & HiF8 & HiF8 - FP16 \\
    \midrule
    DenseNet121       & Top-1 Acc  & 76.04 & $75.93^*$ & $- 0.11$    \\
    EfficientNet-B0   & Top-1 Acc  & 77.33 & $77.08 $  & $- 0.25$    \\
    Inception-V3      & Top-1 Acc  & 77.86 & $77.85^*$ & $- 0.01$    \\
    MobileNet-V2      & Top-1 Acc  & 72.41 & $72.10 $  & $- 0.31$    \\
    ResNet18          & Top-1 Acc  & 70.78 & $70.64 $  & $- 0.14$    \\
    ResNet34          & Top-1 Acc  & 74.41 & $74.25 $  & $- 0.16$    \\
    ResNet50          & Top-1 Acc  & 77.36 & $77.39^*$ & $+ 0.03$    \\
    ResNet101         & Top-1 Acc  & 78.84 & $78.78^*$ & $- 0.06$    \\
    ResNet152         & Top-1 Acc  & 79.42 & $79.41 $  & $- 0.01$    \\
    ResNeXt50         & Top-1 Acc  & 78.14 & $78.01 $  & $- 0.13$    \\
    VGG16             & Top-1 Acc  & 74.07 & $74.02 $  & $- 0.05$    \\
    VGG16-BN          & Top-1 Acc  & 74.59 & $74.54 $  & $- 0.05$    \\
    ViT-Base-Patch16  & Top-1 Acc  & 79.05 & $79.18^*$ & $+ 0.13$    \\
    ViT-Base-Patch32  & Top-1 Acc  & 74.09 & $74.08^*$ & $- 0.01$    \\
    ViT-Large-Patch16 & Top-1 Acc  & 76.08 & $76.45^*$ & $+ 0.37$    \\
    ViT-Large-Patch32 & Top-1 Acc  & 71.77 & $71.82^*$ & $+ 0.05$    \\
    Transformer-Base  & BLEU       & 25.92 & $26.04 $  & $+ 0.12$    \\
    Bert-Large-MRPC   & F1         & 89.67 & $90.02 $  & $+ 0.35$    \\
    YoLo-V3           & mAP50-95   & 43.70 & $43.60 $  & $- 0.10$    \\
    YoLo-V3-Tiny      & mAP50-95   & 16.63 & $16.69^*$ & $+ 0.06$    \\
    DeepLab-V3        & mIOU       & 78.65 & $78.51^*$ & $- 0.14$    \\
    \bottomrule
    \multicolumn{4}{l}{$^*$: Backward rounding for HiF8 is HR, otherwise TA}
  \end{tabular}
\end{table}

Experimental results for the traditional neural networks are listed in 
Table \ref{TNN-Training}. 
Models from DenseNet to ViT series were all trained and evaluated on 
ImageNet ILSVRC2012 dataset. 
Transformer-Base was trained on WMT 2016 English -> German dataset and 
tested on newstest2014 dataset. 
Bert-Large was trained and evaluated on GELU MRPC dataset. 
YoLo series were trained and tested on COCO2017 dataset.
And DeepLab-V3 was trained and evaluated on PASCAL VOC2012 dataset. 
For all metrics in Table \ref{TNN-Training}, higher scores are better. 
To smooth the jitter and deviation caused by weight initialization, the 
validation accuracy of some models was the average value of 2 to 4 rounds 
of training for both FP16 and HiF8. 
As for the validation accuracy of training using TA rounding and hybrid 
rounding, we only presented the results with higher accuracy. 
However, only Yolo-V3-Tiny shows a significant difference in the 
validation accuracy of the two rounding methods. 

From Table \ref{TNN-Training}, we can see that for the CNN structure, 
HiF8 shows slightly lower training accuray than FP16. 
For the Transformer structure, HiF8 shows slightly higher training 
accuracy than FP16. 
In general, for the traditional neural networks, we conclude that 
HiF8 training results with backward global loss-scaling strategy, 
match those of FP16 training sessions. 
This is better than the FP8 training method, which requires per-tensor 
scaling due to the narrow dynamic range \cite{Micikevicius2022}.

\subsection{Inference with Per-Tensor Scaling} 
\label{TNN-PTS}

The model trained by HiF8 can be directly used for inference. 
Therefore, this section only focuses on the process of converting 
high-precision trained model into HiF8 inference model. 
To reflect the capability of HiF8, two inference results of post-training 
quantization (PTQ) will be presented. 
First, we directly cast high-precision activation and weight tensors of 
all layers into HiF8 format without calibration, and compare its accuray 
with the original model. 
Second, we evaluate HiF8 calibration of models trained in FP32 or FP16. 
Unlike the int8 and FP8 calibration, which usually use per-tensor 
scaling for activations and per-channel scaling for weights 
\cite{Micikevicius2022, Anderson2021}, 
in this paper, we only perform the per-tensor scaling for both 
activations and weights of all layers. 

\begin{algorithm}
  \caption{HiF8 Calibration with Per-Tensor Scaling}
  \label{PTS_Calib}
  \begin{algorithmic}[1]
    \REQUIRE{Calibration Dataset, High-Precision Model M}
    \ENSURE{HiF8 Quantized Model}

    \FOR{$l$ in $1^{st}$ to $L^{th}$ layer in M}
      \STATE Forward \& Collect high-precision output of layer $l$: 
      $ O^l = A^l \times W^l $
    \ENDFOR

    \STATE Initialize $O_q^0$ with calibration dataset

    \FOR{$l$ in $1^{st}$ to $L^{th}$ layer in M}
      \FOR{$Ea = [-4, 5]$}
        \FOR{$Ew = [-4, 5]$}

        \STATE Scale \& Cast: $A_q^l = $ To\_HiF8\_TA($O_q^{l-1} \times 2^{Ea}$), $W_q^l = $ To\_HiF8\_TA($W^l \times 2^{Ew}$)
        \STATE MatMul \& Restore: $O_q^l = (A_q^l \times W_q^l) \times 2^{-(Ea+Ew)}$
        \STATE Quantization Error: $Err^l = MSE(O_q^l, O^l)$

        \ENDFOR
      \ENDFOR

      \STATE Find $min(Err^l)$ of all search space \& Store the corresponding $Ea$, and $Ew$
      \STATE Vector operations (Non-linearities \& Normalizations): $O_q^l = Vector(O_q^l)$ 

    \ENDFOR
  \end{algorithmic}  
\end{algorithm}

As shown in Algorithm \ref{PTS_Calib}, we restrict the value of 
each tensor's scaling factor to an integer power of two. 
In this way, there are only a limited number of proper choices for 
scaling factors near $2^0$, and the scaling operations involve 
only addition and subtraction of exponents, no multiplication. 
To find two suitable scaling factors for a tensor multiplication 
inputs, we search for several combinations of scaling factors and 
choose the result that minimizes the MSE (mean squared error) of 
the output tensor. 

\begin{table}[htbp]
  \caption{Validation Accuracy after HiF8 PTQ of Traditional Neural Networks}
  \label{TNN-Inference}
  \centering
  \begin{tabular}{llccccc}
    \toprule 
    Model & Metric & Baseline & HiF8\_Cast & Cast - Baseline & HiF8\_PTS & PTS - Baseline\\
    \midrule
    ResNet18            & Top-1 Acc  & 69.76   & 68.98 & $-0.78$               & 69.48 & \color{blue}{$-0.28$} \\
    ResNet34            & Top-1 Acc  & 73.31   & 72.56 & $-0.76$               & 72.96 & \color{blue}{$-0.35$} \\
    ResNet50            & Top-1 Acc  & 76.13   & 74.85 & $-1.28$               & 75.44 & $-0.69$ \\
    ResNet101           & Top-1 Acc  & 77.37   & 76.48 & $-0.89$               & 76.89 & \color{blue}{$-0.48$} \\
    ResNet152           & Top-1 Acc  & 78.31   & 77.76 & $-0.55$               & 77.96 & \color{blue}{$-0.35$} \\
    ResNeXt50           & Top-1 Acc  & 77.62   & 76.74 & $-0.88$               & 77.17 & \color{blue}{$-0.44$} \\
    VGG16               & Top-1 Acc  & 71.59   & 70.98 & $-0.61$               & 71.28 & \color{blue}{$-0.31$} \\
    DenseNet121         & Top-1 Acc  & 74.43   & 73.58 & $-0.86$               & 74.14 & \color{blue}{$-0.29$} \\
    Inception-V3        & Top-1 Acc  & 77.92   & 76.67 & $-1.25$               & 77.38 & $-0.54$ \\
    ViT-Base-Patch16    & Top-1 Acc  & 81.07   & 80.86 & \color{blue}{$-0.21$} & 80.91 & \color{blue}{$-0.15$} \\
    ViT-Base-Patch32    & Top-1 Acc  & 75.92   & 75.66 & \color{blue}{$-0.26$} & 75.85 & \color{blue}{$-0.07$} \\
    ViT-Large-Patch16   & Top-1 Acc  & 79.68   & 79.69 & \color{blue}{$+0.00$} & 79.77 & \color{blue}{$+0.09$} \\
    ViT-Large-Patch32   & Top-1 Acc  & 76.96   & 76.83 & \color{blue}{$-0.13$} & 76.87 & \color{blue}{$-0.09$} \\
    MaskRCNN            & bbox       & 37.8    & 37.1  & $-0.7 $               & 37.3  & \color{blue}{$-0.5 $} \\
    SSD-VGG16           & bbox       & 25.1    & 24.2  & $-0.9 $               & 24.8  & \color{blue}{$-0.3 $} \\
    YoLo-V3             & mAP50-95   &$ 43.3^*$& 42.2  & $-1.1 $               & 42.8  & \color{blue}{$-0.5 $} \\
    3D-Unet             & mean-dice  &$90.98^*$& 91.03 & \color{blue}{$+0.05$} & /     & /                     \\
    Bert-Large-MRPC     & F1         &$87.44^*$& 87.19 & \color{blue}{$-0.25$} & 87.63 & \color{blue}{$+0.19$} \\
    Bert-Large-SQuADv1.1& F1         &$91.40^*$& 91.39 & \color{blue}{$-0.01$} & /     & /                     \\
    \bottomrule
    \multicolumn{7}{l}{$^*$: Baseline was trained in FP16, otherwise FP32}         \\
    \multicolumn{7}{l}{HiF8\_Cast: Directly cast F32/16 tensors to HiF8}           \\
    \multicolumn{7}{l}{HiF8\_PTS: Calibration with Per-tensor Scaling}             \\
  \end{tabular}
\end{table}

Table \ref{TNN-Inference} lists the inference results of HiF8 for 
the traditional neural networks. 
In addition to some of the previously mentioned models, we further 
evaluated the inference accuracy of MaskRCNN \cite{He2017} and 
SSD-VGG16 \cite{Liu2016} in detection applications, as well as the 
3D-Unet \cite{Cicek2016} in segmentation scenarios. 
For all metrics in Table \ref{TNN-Inference}, higher scores are better. 
We define the metric loss of no greater than 0.5 as the ideal inference 
result, and mark it blue. 
It can be seen that Transformer-based models can be used for inference 
after direct conversion to HiF8 without calibration. 
When per-tensor scaling is enabled for both activations and weights, 
most CNNs can be used for inference. 

To improve HiF8 inference accuracy for the traditional neural networks, 
two further methods can be tried. 
First, based on per-tensor scaling for both activations and weights, 
we can leave some sensitive layers as FP16 or BF16. 
For example, if we convert the first layer of ResNet50 and Inception-V3 
to FP16 and the other layers to HiF8, their accuracy losses improve to 
be $0.20$ and $0.23$. 
Second, like int8 and FP8, we can use per-tensor scaling for activations 
and per-channel scaling for weights.

\section{Experiments on Large Language Models} 

LLMs exhibit some unique features that differ from the traditional 
neural networks. 
On the training side, the gradients distribution is more dispersed. 
Thus larger dynamic range or special technique, is required to avoid 
too much data becoming zero during data type conversion\cite{Choquette2022} . 
In terms of inference, outliers play an important role in validation 
accuracy. 
Therefore, dedicated method such as SmoothQuant \cite{Xiao2023}, is 
possible needed to reduce the quantization error of outliers. 
In this section, to apply HiF8 on LLMs, we proposed and tried three 
training methods, each with different costs and coverage. 
We also evaluated three inference methods, including direct-cast, 
per-tensor scaling, and SmoothQuant. 
Note that for LLMs, only TA rounding was used for HiF8.

\subsection{Training}

First, we inroduce the HiF8 training methods for LLMs in our experiments: 
\begin{itemize}
  
  \item \textbf{Backward Loss-Scaling (BLS)}
  
  This method is inherited from FP16 mixed-precision training 
  \cite{Micikevicius2017}. 
  More details can be found in Section \ref{BLS}. 
  
  \item \textbf{Adaptive Loss-Scaling (ALS)}
  
  This is an optimal configuration for the backward loss-scaling. 
  In the current mixed-precision training, scale window (growth\_interval) 
  is an input constant throughout the training task
  \footnote{\url{https://pytorch.org/docs/stable/_modules/torch/cuda/amp/grad_scaler.html}}, 
  and is usually set to 1000 or 2000 for the traditional neural networks. 
  In some LLMs, the maginitude of gradients changes rapidly in 
  the early iterations. 
  At this moment, large scale window cannot catch up with the change 
  in gradients, resulting in a failure of convergence. 
  However, a scale window that is too small can significantly 
  degrade the performance of training. 

  To address this dilemma, we propose the adaptive loss-scaling. 
  Specifically, we first define an incremental list of scale window, 
  such as \{1, 20, 50, 100, 200, 500, 1000\}. 
  Then the inital scale value and scale window are set to $2^{32}$ and 20. 
  During training, if the scale value increases three times, the 
  scale window will increase once as per the list order. 
  If the scale value decreases three times in a row, scale window 
  will decrease once as per the list order. 
  The proposed ALS can be applied to both FP16 and HiF8 to improve 
  the stability of training. 
  Please refer to the website for more information 
  \footnote{\url{https://mindformers.readthedocs.io/zh-cn/latest/docs/feature_cards/Training_Algorithms.html}}. 

  \item \textbf{Per-Tensor Scaling (PTS)}
  
  Although training with BLS and ALS can be successfully carried out in HiF8 for 
  a number of LLMs, there are cases where per-tensor scaling is needed to 
  improve accuracy and mitigate the difficulty of hyper-parameters tuning. 
  To be specific, we define a scale factor for each GEMM input tensor, including 
  activation, weight, and activation gradient. 
  The initial scale factors are all set to 1. 
  Then every 10 iterations, we compute the maximun absolute value (Amax) of each 
  tensor, and update the corresponding scale factor by a certain algorithm. 
  To avoid additional rounding errors, the scale factors are restricted to integer 
  powers of 2. 
  While in each iteration, whether the scale factors are updated or not, they 
  will scale the corresponding tensors to a better range that can be represented 
  by HiF8. 
  Finally, GEMM outputs need to be descaled to restore the correct results. 
  Note that scale and descale operations are all carried out in high-precision 
  formats, such as FP32, BF16, and FP16 (with BLS or ALS). 

  Actually, the PTS we used for HiF8 is very similar to the transformer 
  engine for FP8 \cite{Choquette2022}.
  However, because HiF8 has much larger dynamic range than FP8 (especially E4M3), 
  we do not need to compute Amax in all iterations, thereby greatly reducing 
  the occupation of vector resources. 
  
\end{itemize}

In the training experiments for LLMs, we selceted T5 \cite{Raffel2020}, LLaMA 
\cite{Touvron2023}, and GPT3 \cite{Brown2020} to evidence the capability of HiF8. 
Our training datasets are several mixtures of three sources, including Book3 
(101 GB) and OpenWebText2 (63 GB) from the Pile dataset \cite{Gao2021}, and 
Wikipedia (20 GB). 

\begin{figure}[htbp]
  \centering
  \includegraphics[width=17.75cm]{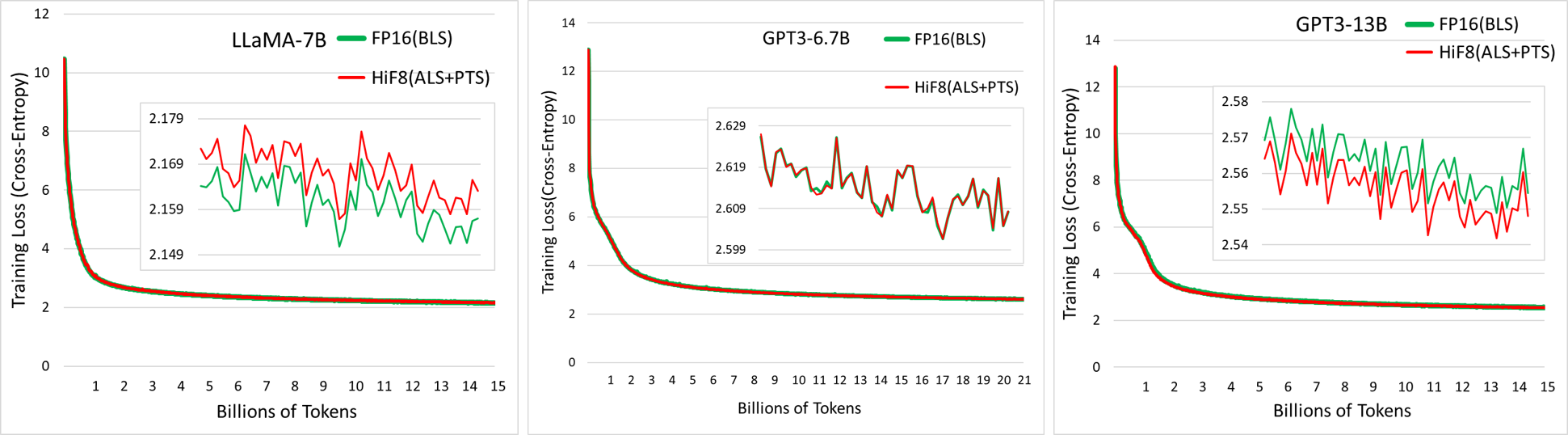}
  \caption{Training Loss Curves for Large Language Models}
  \label{LLM-Curve}
\end{figure}

In Fig. \ref{LLM-Curve}, the training loss (average of every 10 iterations) over 
tokens is displayed for LLaMA-7B and GPT3 models of 6.7B and 13B parameters. 
The weight initializations (non-fixed random seeds) and optimizer hyper-parameters 
are consistent across models trianed with HiF8 and FP16. 
Again, only GEMM inputs, including activation, weight, and activation gradient 
tensors, are casted from FP16 to HiF8. 
As shown in Fig. \ref{LLM-Curve}, from a global view, the loss curves overlap 
with each other very well. 
And from a locally enlarged view, we can see that the difference between HiF8 
and FP16 is within run-to-run variation caused by different random seeds. 

Validation perplexities (PPL: lower is better) for a variety of LLMs are 
listed in Table \ref{LLM-Train}. 
FP16 baselines are all trained with BLS strategy. 
From Table \ref{LLM-Train}, we can conclude that HiF8 traning results for LLMs, 
match those of FP16 training sessions. 
Specifically, with very small penalty of training accuray, HiF8 with BLS or ALS 
has almost no extra overhead compared to the FP16 baseline.  
While ALS technique can obviously mitigate the difficulty of hyper-parameters 
tuning, such as the ratio of warmup iterations. 

\begin{table}[htbp]
  \caption{Validation Perplexity for Large Language Models}
  \label{LLM-Train}
  \centering
  \begin{tabular}{llclccc}
    \toprule 
    Model     & Training Data    & Training Tokens & HiF8 Strategy & FP16  & HiF8  & HiF8 - FP16 \\
    \midrule
    T5-11B    & Wiki                 & 3.28B           & BLS           & 7.40  & 7.53  & $+0.13$ \\
    GPT3-350M & Wiki, OpenWeb, Book3 & 14.8B           & BLS           & 17.43 & 17.52 & $+0.09$ \\
    GPT3-2.7B & Wiki, OpenWeb, Book3 & 10.5B           & BLS           & 14.96 & 15.04 & $+0.08$ \\
    GPT3-6.7B & Wiki, OpenWeb, Book3 & 21.3B           & ALS           & 13.06 & 13.14 & $+0.08$ \\
    GPT3-6.7B & Wiki, OpenWeb, Book3 & 21.3B           & ALS + PTS     & 13.06 & 12.99 & $-0.07$ \\
    GPT3-13B  & Wiki, OpenWeb, Book3 & 14.8B           & ALS + PTS     & 12.62 & 12.54 & $-0.08$ \\
    LLaMA-7B  & Wiki, OpenWeb        & 14.8B           & ALS + PTS     & 8.25  & 8.28  & $+0.03$ \\
    LLaMA-13B & Wiki, OpenWeb        & 8.85B           & ALS + PTS     & 8.71  & 8.74  & $+0.03$ \\
    \bottomrule
  \end{tabular}
\end{table}

Meanwhile, with the extra computation of Amax every 10 iterations, 
HiF8 with PTS strategy sometimes yields better training accuracy 
than the FP16 baseline. 
Fortunately, thanks to the large daynamic range of HiF8, such additional overhaed 
is much smaller than the FP8 with transformer engine \cite{Choquette2022}, 
resulting in better training performance on some neural networks.

\subsection{Inference}

The LLM trained by HiF8 can be directly used for inference. 
Thus we only consider the HiF8 PTQ of LLMs trained in higher precision. 
In addition to the PTQ methods of direct-cast and per-tensor scaling used in 
Section \ref{TNN-PTS}, 
we further evaluated the SmoothQuant \cite{Xiao2023} calibration method for LLMs, 
due to the influence of outliers. 
First, to objectively reflect the inference ability of HiF8, 
we chose the quantization-tolerant LLaMA \cite{Touvron2023}, 
and the quantization-sensitive OPT \cite{Zhang2022} as the experimental LLMs.

\begin{table}[htbp]
  \caption{WikiText2 Perplexity after HiF8 PTQ of Large Language Models}
  \label{LLM-Inference-Wiki}
  \centering
  \begin{tabular}{lccccccc}
    \toprule 
    Model     & FP16 & HiF8\_Cast & Cast - FP16   & HiF8\_PTS & PTS - FP16 & HiF8\_SQ & SQ - FP16 \\
    \midrule
    LLaMA-7B  & 5.00 & 5.06  & \color{blue}{0.06} & 5.02 & \color{blue}{0.02} & 5.02 & \color{blue}{0.02} \\
    LLaMA-13B & 4.54 & 4.60  & \color{blue}{0.06} & 4.55 & \color{blue}{0.01} & 4.58 & \color{blue}{0.04} \\
    LLaMA-65B & 3.13 & 3.16  & \color{blue}{0.03} & 3.16 & \color{blue}{0.03} & 3.16 & \color{blue}{0.03} \\
    OPT-7B    & 9.08 & 10.68 & \color{red}{1.60}  & 9.54 & 0.46               & 9.38 & \color{blue}{0.30} \\
    OPT-13B   & 8.64 & 9.11  & 0.47               & 9.08 & 0.44               & 8.72 & \color{blue}{0.08} \\
    OPT-66B   & 7.74 & 106.9 & \color{red}{99.16} & 8.16 & 0.42               & 7.92 & \color{blue}{0.18} \\
    \bottomrule
    \multicolumn{8}{l}{HiF8\_Cast: Directly cast F32/16 tensors to HiF8}   \\
    \multicolumn{8}{l}{HiF8\_PTS: Calibration with Per-tensor Scaling}     \\
    \multicolumn{8}{l}{HiF8\_SQ: Calibration with SmoothQuant}             \\
  \end{tabular}
\end{table}

As shown in Table \ref{LLM-Inference-Wiki}, for some quantization-tolerant 
LLMs such as LLaMA, HiF8 can easily achieve the desired inference accuracy 
even if only the simplest direct-cast is performed. 
But for some quantization-sensitive LLMs like OPT, dedicated calibration 
methods, such as PTS and SmoothQuant, are needed to improve the inference 
accuray of HiF8. 
More PTS calibration results for downstream tasks are outlined in 
Table \ref{LLM-Inference-PTS}. 

\begin{table}[htbp]
  \caption{Inference Accuracy after HiF8 PTS Calibration for Large Language Models}
  \label{LLM-Inference-PTS}
  \centering
  \begin{tabular}{lllccc}
    \toprule 
    Model     & Downstream Task & Metric                        & FP16  & HiF8  & Metric Loss \\
    \midrule 
    GPT3-2.7B & MMLU            & Acc\_Avg (5-shots) $\uparrow$ & 27.73 & 27.90 & \color{blue}{$-0.17$} \\
    GPT3-2.7B & WikiText103     & PPL $\downarrow$  & 12.05 & 12.20 & \color{blue}{$+0.15$} \\
    LLaMA-7B  & Lambada         & Acc (zero-shot) $\uparrow$    & 89.37 & 89.21 & \color{blue}{$+0.16$} \\
    \bottomrule
  \end{tabular}
\end{table}

Finally, as described in Section \ref{TNN-PTS}, per-tensor scaling 
for activations and per-channel scaling for weights, and some combination 
of different calibration methods, are worth trying to further improve 
the HiF8 inference accuracy of LLMs.

\section{Conclusion and Outlook}

In this preliminary white paper, we propose a novel 8-bit floating-point format 
HiFloat8, consisting of sign, dot, exponent, and mantissa fields. 
By deeply exploring the match between format and data distribution, HiF8 achieves 
a much better balance between precision and daynamic range than the existing 8-bit 
formats. 
Experments on a large number of traditional neural networks and LLMs, demonstrate 
that as a single format, HiF8 works well in both training and inference. 
In the future, we will disclose another research achievement of HiFloat project: 
HiFloat below 8-bit, as well as its training and inference capabilities.


\bibliographystyle{ieeetr}
\bibliography{neurips_2023}

\end{document}